\documentclass{article}

\usepackage{amsmath,amssymb,amsfonts,amstext}%
\usepackage{graphicx}%
\usepackage{multirow}%
\usepackage{import}
\usepackage{algorithm}%
\usepackage{algorithmicx}%
\usepackage{algpseudocode}%
\usepackage[preprint, nonatbib]{neurips_2024}

\usepackage[utf8]{inputenc} % allow utf-8 input
\usepackage[T1]{fontenc}    % use 8-bit T1 fonts
\usepackage{hyperref}       % hyperlinks
\usepackage{url}            % simple URL typesetting
\usepackage{booktabs}       % professional-quality tables
\usepackage{nicefrac}       % compact symbols for 1/2, etc.
\usepackage{microtype}      % microtypography
\usepackage{xcolor}         % colors
\usepackage{biblatex} %Imports biblatex package

\title{TeVAE: A Variational Autoencoder Approach for Discrete Online Anomaly Detection in Variable-state Multivariate Time-series Data}

\author{%
  Lucas Correia\\
  Mercedes-Benz AG\\
  Stuttgart\\
  Germany \\
  \texttt{lucas.correia@mercedes-benz.com} \\
  \And
  Jan-Christoph Goos \\
  Mercedes-Benz AG\\
  Stuttgart\\
  Germany \\
  \AND
  Philipp Klein \\
  Mercedes-Benz AG\\
  Stuttgart\\
  Germany \\
  \And
  Thomas Bäck \\
  Leiden University \\
  Leiden \\
  The Netherlands \\
  \And
  Anna V. Kononova \\
  Leiden University \\
  Leiden \\
  The Netherlands \\
}

\newcommand{\lc}[1]{\textcolor{black}{#1}}

\algnewcommand\algorithmicinput{\textbf{Input:}}
\algnewcommand\Input{\item[\algorithmicinput]}

\algnewcommand\algorithmicresult{\textbf{Result:}}
\algnewcommand\Result{\item[\algorithmicresult]}

\begin{document}

\maketitle

\begin{abstract}
As attention to recorded data grows in the realm of automotive testing and manual evaluation reaches its limits, there is a growing need for automatic online anomaly detection. 
This real-world data is complex in many ways and requires the modelling of testee behaviour. 
To address this, we propose a temporal variational autoencoder (TeVAE) that can detect anomalies with minimal false positives when trained on unlabelled data. 
Our approach also avoids the bypass phenomenon and introduces a new method to remap individual windows to a continuous time series. 
Furthermore, we propose metrics to evaluate the detection delay and root-cause capability of our approach and present results from experiments on a real-world industrial data set. 
When properly configured, TeVAE flags anomalies only 6\% of the time wrongly and detects 65\% of anomalies present. 
It also has the potential to perform well with a smaller training and validation subset but requires a more sophisticated threshold estimation method.
\end{abstract}

\section{Introduction}\label{sec:introduction}
Anomaly detection in time-series data has emerged as a common problem with a variety of real-world applications, from the medical field \cite{moody_impact_2001} to high-performance computing \cite{su_robust_2019}.
Findings from the field of anomaly detection are also of particular interest to the modern automotive industry.
This industry is very diverse, including, for example, manufacturing, prototyping and testing.
Furthermore, a car is a complex structure that can be segmented into different subsystems, one of them being the powertrain, which includes all components required for longitudinal dynamics.
The powertrain is an important subsystem, as it is something a customer interacts with when driving or being driven.
Therefore, testing the powertrain is an an integral part of the wider automotive powertrain development and is undertaken at different stages of development. 
Each of these stages is composed of many integration levels. These integration levels range from powertrain sub-component testing, such as the electric drive unit (EDU) controller or high-voltage battery (HVB) management system, to whole vehicle powertrain testing. 
For each integration level there is a special type of controlled environment, called a test bench. The use-case in this paper is on an endurance powertrain test bench, where the EDU and HVB on their own are tested under different conditions and loads for longer periods to simulate wear over time. 
Given the costly maintenance and upkeep costs of such test benches, it is desirable to keep downtime at a minimum and to avoid faulty records. 
Also, it is desirable to detect problems early to prevent damage to the testee. 
Online time-series anomaly detection is especially relevant, as it can provide timely insights into potentially harming behaviour that deviates from the norm.

Applying anomaly detection to this real-world use case is especially challenging due to its complexity and highly dynamic setup.
During powertrain testing several sensors record signals over time, leading to data in the form of multivariate time series. 
These signals not only have correlations within themselves over time but also between each other.
In addition to that, some of the signals recorded also feature variable-state behaviour, meaning the same test done at different times will yield slightly different data.
This is due to certain channels like the battery signals presenting slightly different behaviour depending on how warm or charged the battery is.
All these characteristics exclude the use of simpler statistical models as they are not compatible with all of the mentioned data properties.

Given that evaluation is currently done manually by inspection, it is not feasible to analyse every single test, also evaluation tends to be delayed, only being undertaken days after the test is recorded, hence there is a clear need for automatic, fast and unsupervised evaluation methodology which can flag anomalous behaviour before the next testing procedure is started.

To achieve this, we propose a temporal multi-head attention-based variational autoencoder (TeVAE).
TeVAE consists of a bidirectional long short-term memory (BiLSTM) variational autoencoder architecture that maps a time-series window into a temporal latent distribution \cite{park_multimodal_2018} \cite{su_robust_2019}. 
Also, a multi-head attention (MA) mechanism is added to further enhance the sampled latent matrix before it is passed on to the decoder. 
As shown in the ablation study, this approach avoids the so-called bypassed phenomenon \cite{bahuleyan_variational_2018}, which is the first contribution. 
Furthermore, this paper offers a unique methodology for the reverse-window process.
It is used for remapping the fixed-length windows the model is trained on to continuous variable-length sequences, i.e.\ time series.
Moreover, we propose a set of metrics apt to \textit{online} time-series anomaly detection that is not only interpretable and simple but is also compatible with discrete time-series anomaly detection.
Lastly, the root-cause capability of TeVAE is investigated, for which a new metric, the root-cause precision, is also proposed.

This paper is structured as follows: First, a short background is provided in Section \ref{sec:background} on the powertrain testing methodology specific to this use case, as well as the theory behind VAEs and MA mechanisms. 
Then, related work in variational autoencoder-based time-series anomaly detection is presented in Section \ref{sec:related_work}, followed by an in-depth introduction of the real-world data set and the approach we propose in Section \ref{sec:proposed_approach}. 
Then, several experiments testing different aspects of the proposed method are conducted and discussed in Section \ref{sec:results}, along with the final results. 
Finally, conclusions from this work are drawn and an outlook into future work is provided in Section \ref{sec:conclusion}. 
The source code for the data pre-processing, model training as well as evaluation can be found under \url{https://github.com/lcs-crr/TeVAE}.

\section{Background}\label{sec:background}
\subsection{Real-world Application: Automotive Powertrain Testing}
During endurance testing a portfolio of different drive cycles is run, where a drive cycle is a standardised driving pattern characterised by the vehicle speed, which enables repeatability.
For this type of testing, the portfolio consists exclusively of proprietary drive cycles, which differ from the public drive cycles used, for example, for vehicle fuel/energy consumption certification. 
The reason why proprietary drive cycles are used for endurance runs is that they allow for more extensive loading of the powertrain. 

Given the presence of a battery in the testee, some time has to be dedicated to battery soaking (sitting idle) and charging. 
These procedures are also standardised using soaking and charging cycles, respectively, although, for the intents and purposes of this paper, they are omitted. 
\lc{What is left in the portfolio} are eight dynamic drive cycles representing short, long, fast, slow and dynamic trips ranging from $5$ to $30$ minutes. 
\lc{Modelling the testee behaviour is further complicated through variable-state behaviour.
This means that two measurements of the same drive cycle done one after another will look different, depending on initial states. A measurement is defined as an instance of a drive cycle and is in the form of a multivariate time series or sequence.
Variable-state behaviour can be categorised into short-term reversible and long-term irreversible. 
On the one hand, there are channels like the battery temperature or state of charge (SoC) that contribute to the short-term reversible kind since the battery heats up and discharges as it is used.
On the other hand, processes like battery ageing, also known as the state of health (SoH), contribute to long-term irreversible behaviour, which is not considered and modelled in this use-case.}

On powertrain test benches, there are several control methods to ensure the testee maintains the given drive cycle, i.e.\ the vehicle speed profile. 
In this particular test bench, the regulation is done by the acceleration pedal \lc{position} and the EDU revolutions per minute (rpm).

\subsection{Data Set}
\lc{To enable the development of anomaly detection methodology that can deal with the above-mentioned challenges, a data set is created using the data for one of the testees.}
This real-world data set $\mathcal{D}$ consists of \lc{thousands} measurement files, each \lc{of variable length and containing} hundreds of (many redundant or empty) channels. 
\lc{The training subset $\mathcal{D}^{\text{train}}$ consists of $M=2785$ unlabelled time series such that $\mathcal{D}^{\text{train}} = [\mathcal{S}_1, ..., \mathcal{S}_m, \newline ..., \mathcal{S}_{M}]$. 
Note that each time series in $\mathcal{D}^{\text{train}}$ has variable length $T_{m}$ and dimensionality $d_\mathcal{D}$, such that $\mathcal{S}_m \in \mathbb{R}^{T_{m} \times d_\mathcal{D}}$.}
For this work, a list of \lc{representative} $d_{\mathcal{D}}=13$ channels is hand-picked in consultation with the test bench engineers. 
The chosen features along with their indices \lc{are as shown in Table \ref{tab:legend}.
\begin{table}[!t]
    \caption{Channels (features) chosen for modelling testee behaviour \cite{correia_ma-vae_2023}. Indices correspond to Figure \ref{fig:data_plot}.}
    \label{tab:legend}
    \centering
    \begin{tabular}{cll}
    Index   & Name                      \\ \hline\hline
    1       & Vehicle Speed             \\
    2       & EDU Torque                \\
    3       & Left Axle Torque          \\
    4       & Right Axle Torque         \\
    5       & EDU Current               \\
    6       & EDU Voltage               \\
    7       & HVB Current               \\
    8       & HVB Voltage               \\
    9       & HVB Temperature           \\
    10      & HVB State of Charge       \\
    11      & EDU Rotor Temperature     \\
    12      & EDU Stator Temperature    \\
    13      & Inverter Temperature      \\ \hline
    \end{tabular}
\end{table}
}

\lc{The testing subset $\mathcal{D}^{\text{test}}$ consists of $N$ labelled time series such that $\mathcal{D}^{\text{test}} = [\mathcal{S}_1, ..., \mathcal{S}_n, ..., \mathcal{S}_{N}]$, where each time series in $\mathcal{D}^{\text{test}}$ has variable length $T_{n}$ and dimensionality $d_\mathcal{D}$, such that $\mathcal{S}_n \in \mathbb{R}^{T_{n} \times d_\mathcal{D}}$.
The labelled anomaly-free portion of the testing subset $\mathcal{D}^{\text{test}}$ accounts for $N_\text{af}=698$ measurements, where a measurement is considered anomaly-free when the testee behaviour conforms to the norm.}

Due to the absence of labelled anomalies in the dataset, realistic anomalous events are intentionally simulated and recorded following the advice of test bench engineers. 
\lc{To this end, four anomaly types are recorded. 
Every anomaly type is recorded for every drive cycle at least once, leading to $N_\text{a}=47$ anomalous measurements that are all used as the labelled anomalous portion of the testing subset $\mathcal{D}^{\text{test}}$. 
Hence $\mathcal{D}^{\text{test}}$ is made up of $N=745$ measurements, representing an anomaly ratio of around $N_\text{a}/N=6.3\%$, however in reality this value is estimated to be much lower.}
This amount of anomalous data in relation to anomaly-free data is used as it approximately matches the anomaly ratio in public data sets \cite{mathur_swat_2016, ahmed_wadi_2017, hundman_detecting_2018, su_robust_2019} and because the data set is not large enough to create a larger anomaly-free test subset. 

\lc{In the first type, the virtual wheel diameter is changed, such that the resulting vehicle speed deviates from the norm. 
The wheel diameter is a parameter as resistances are connected to the shafts rather than actual wheels. 
This accounts for 16 time-series anomalies and the only channel that demonstrates anomalous behaviour is the vehicle speed, since:}
\begin{equation}
    v_{\text{vehicle}}=r \cdot \omega
\end{equation}
where $r$ is the wheel radius and $\omega$ the angular velocity \lc{of the drive shaft}. 
Logically, the anomalous behaviour is most visible at higher speeds.

\begin{figure}[!t]
    \centering
    \includegraphics[width=\textwidth]{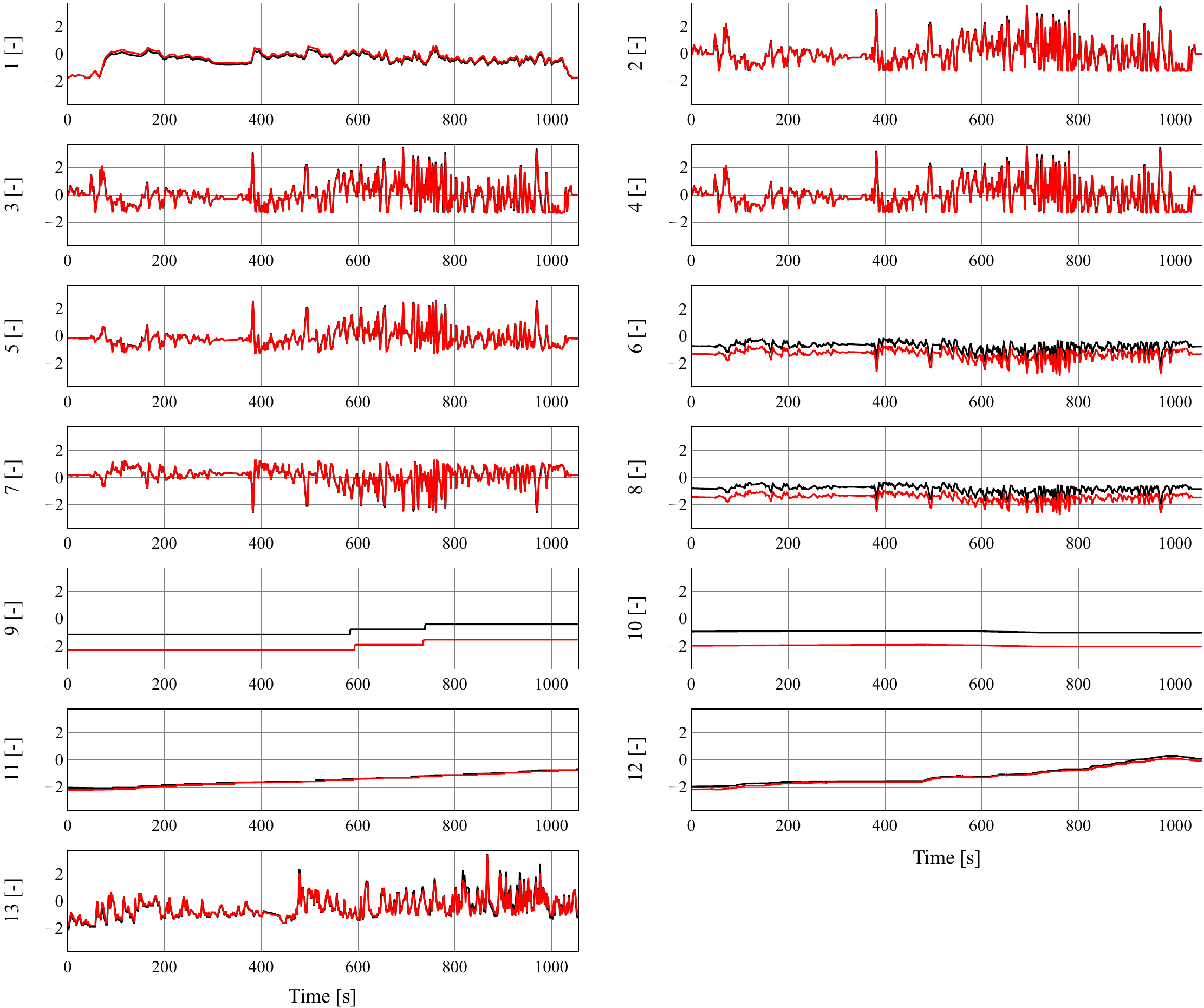}
    \caption{Features of an anomaly-free (black) and an anomalous (red) measurement plotted with respect to time. The anomalous measurement plotted represents a scenario where the wheel diameter has not been set correctly. The amplitude axis is z-score normalised to comply with confidentiality guidelines.}
    \label{fig:data_plot}
\end{figure}

A plot of one anomaly-free (black) and one wheel-diameter anomalous (red) measurement is shown in Figure \ref{fig:data_plot}.
% For the corresponding channel names refer to the legend in Table \ref{tab:legend} for context.
Visual inspection may suggest that the red plot is trivial to detect given the large deviation in the EDU and HVB voltage, temperature and state of charge compared to the black plot.
However, this deviation is in fact normal and to be expected since they depend on how charged the battery is and on how much the battery is used previous to the current measurement. 
\lc{This variable-state behaviour makes anomalies much harder to detect than those in public data sets with such behaviour.}

In the case of the next anomaly type, the recuperation level is turned from maximum to zero, hence the minimum EDU torque is always non-negative and the HVB SoC experiences a higher drop in SoC. 
\lc{Hence, some deviation should be visible in the EDU torque, left and right axle torques and in the HVB state of charge. 
This anomaly type accounts for another eight time-series anomalies.}

For the following anomaly, the HVB is swapped for a battery simulator, where the HVB voltage behaviour deviates from a real battery. 
\lc{Considering that operation works by requesting a given amount of power, a different voltage behaviour will also result in a change in current, since power $P$ is the product of voltage across the battery $V$ and the current $I$:}
\begin{equation}
    P=V\cdot I
\end{equation}
\lc{Therefore, this type of anomaly will be evident in the HVB and EDU voltage channels, as well as the respective current channels, representing 15 time-series anomalies.}

The inverter and EDU share a cooling loop, whose cooling capacity is reduced at the beginning or middle of the measurement \lc{for the last type of anomaly. 
This leads to, for example,} higher EDU rotor, EDU stator and inverter temperatures than normal. 
\lc{For three of the sequences the capacity is reduced at the midpoint of the measurement, whereas another five sequences are recorded with reduced cooling capacity from the very beginning.}

\lc{Clearly this data set consists of several discrete time series.
This is in contrast to \textit{continuous} anomaly detection, which is defined as detecting anomalies in a process that exists for a longer continuous time period without breaks and is the most common type present in public data sets. 
This includes \textit{monitoring} applications like in water distribution \cite{mathur_swat_2016, ahmed_wadi_2017} or server machines \cite{su_robust_2019}.
Use cases of continuous anomaly detection tend to consist of a singular longer time series which contains anomaly-free and anomalous sub-sequences within it.
On the other hand, we define \textit{discrete} anomaly detection as detecting anomalies in chunks of processes that happen independently of each other, such as automotive test benches, where several tests may occur one after another but are not temporally contiguous and hence provide a multivariate time series for each test. 
Here the testees are not monitored over a longer period of time, but rather the measured time series are \textit{evaluated} as they are being recorded. 
Arguably, analysis of heartbeat rhythms could be considered a discrete anomaly detection problem if applied generically to a number of test subjects, rather than individually, as the training and testing subsets each likely consist of multiple time series from different subjects.
Therefore, data sets for discrete anomaly detection consist of several anomaly-free and anomalous time series, where a given anomalous time series may be entirely anomalous or only partly.}

Given that some channels (such as torque) are sampled much faster than others (like temperature and SoC), a common sampling rate of $2$Hz is chosen. 
Channels sampled slower than $2$Hz are linearly interpolated, which is considered permissible due to the lower amplitude resolution of those channels.
Channels sampled faster than $2$Hz are passed through a low-pass filter with a cut-off frequency of $1$Hz and then resampled to $2$Hz, as is consistent with the Whittaker–Nyquist–Shannon theorem \cite{shannon_communication_1949}.
Then, the measurements are z-score normalised, i.e.\ transformed such that the mean for each channel lies at $0$ and the standard deviation at $1$.
Lastly, the measurements are windowed to create a set of fixed-length sub-sequences, or \textit{windows}. 
\lc{The reasoning behind using windows rather than entire sequences is that most of the dynamic processes in the time-series data are fairly fast and hence only have short-term effect that lasts for a small period of time. 
Modelling entire variable-length sequences would be possible but much of the model learning capacity would be wasted on internal state mechanisms carrying information for longer periods of time than necessary. 
This therefore allows for more efficient model training on windows that are only as long as they need to be to represent all present dynamics in the time-series data.
To find the effect length of the slowest dynamics an autocorrelation analysis is undertaken for each of the drive cycles and each of the features within them.
This is in contrast to approaches taken in literature, which treat window size as a hyperparameter that is tuned.
It is unclear how hyperparameter tuning can be undertaken outside of a supervised problem, hence it is assumed that in the real world, this process would not be possible.
An autocorrelation analysis yields the correlation between the time series and a range of lagged versions of the time series for a specified number of lags and does not require labels. 
An example of an autocorrelation plot for an arbitrary measurement and arbitrary channel is shown in Figure \ref{fig:autocorrelation}.
\begin{figure}[b!]
    \centering
    \includegraphics[width=0.8\textwidth]{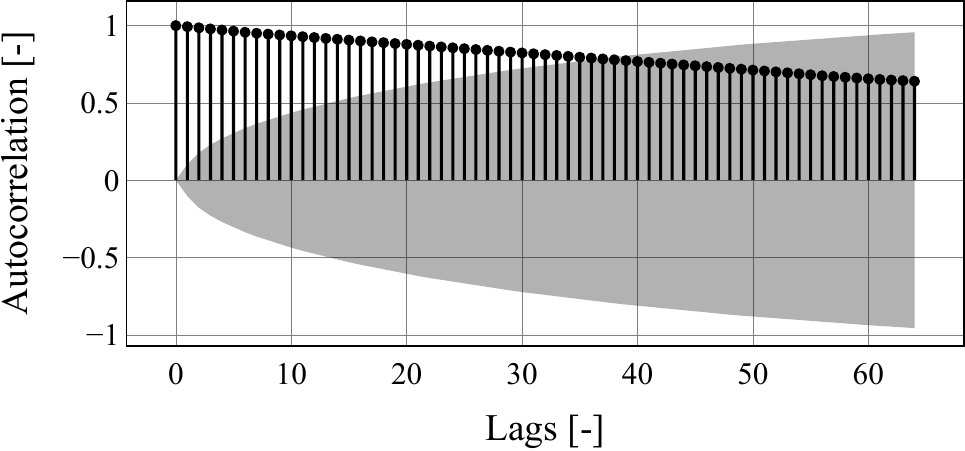}
    \caption{\lc{Example plot for the autocorrelation as a function of lags. The band shown represents the confidence interval, below which we assume the autocorrelation is no longer statistically significant.}}\label{fig:autocorrelation}
\end{figure}
As is evident, the autocorrelation decreases with the number of lags, which is consistent with theory as dynamic effects dampen with time.
Therefore, the slowest dynamics present that is still significant is at the lag where the autocorrelation intersects the confidence interval.
The window size is therefore the slowest dynamic present in any channel in all sequences.
For computational efficiency, the window size $w$ is then set as the smallest power of two larger than the largest lag, in this case, $w=256$ time steps or $128$ seconds.}
Each window overlaps its preceding and succeeding windows by half a window, i.e.\ the shift between windows is $w/2=128$ time steps, in order to reduce computational load compared to a shift of one time step.

In an operative environment, it is desirable to find out whether the previously recorded sequence had any problems to analyse before the next measurement is recorded.
Also, a model that performs as well as possible with as little data as possible translates to faster deployment.
Good performance is indicated by a model that can detect as many anomalies as possible and rarely labels anomaly-free measurements wrongly.
To investigate the required training subset size of the model, it is trained with $1$h, $8$h, $64$h, and $512$h of dynamic testing time, which corresponds to the first $6$, $44$, $348$, and $2785$ measurements, respectively.
The results are also represented in Section \ref{sec:results}. 
In each of the above-mentioned cases, the \lc{unlabelled} training subset \lc{$\mathcal{D}^{\text{train}}$} is further split into a training ($80\%$) and a validation ($20\%$) subsets.

\subsection{Variational Autoencoders}
The variational autoencoder \cite{kingma_auto-encoding_2014}\cite{rezende_stochastic_2014} is a generative model that structurally resembles an autoencoder, but is theoretically derived from variational Bayesian statistics.
As opposed to the regular deterministic autoencoder, the VAE uses the evidence lower bound (ELBO), which is a lower bound approximation of the so-called log evidence $\log p_\theta(\textbf{X})$, as its objective function.
The ELBO, Equation \ref{eq:ELBO}, can be expressed as the reconstruction log-likelihood and the negative Kullback-Leibler Divergence ($D_{\text{KL}}$) between the approximate posterior $q_{\phi}(\textbf{Z}\vert \textbf{X})$ and the prior $p_{\theta}(\textbf{Z})$, which is typically assumed to be a Gaussian distribution \cite{goodfellow_deep_2016}. 
\lc{The training data does not have to follow a Gaussian distribution since the latent distribution is a nonlinear mapping of the training data.}
\begin{equation}
    \begin{split}
        \mathcal{L}_{\theta, \phi}(\textbf{X}) &= \mathbb{E}_{\textbf{Z} \sim q_\phi(\textbf{Z}|\textbf{X})}  \left[ \log p_\theta(\textbf{X} | \textbf{Z}) \right] \\
        & - D_{\text{KL}}(q_\phi(\textbf{Z} | \textbf{X}) || p_\theta(\textbf{Z}))  \\
    \end{split}
    \label{eq:ELBO}
\end{equation}
where $\textbf{Z} \in \mathbb{R}^{w \times d_{\textbf{Z}}}$ is the sampled latent matrix and \lc{$\textbf{X} \in \mathbb{R}^{w \times d_{\mathcal{D}}}$} is the input window.
$w$ refers to the window length, whereas \lc{$d_{\mathcal{D}}$} and $d_{\textbf{Z}}$ refer to the \lc{data set} and latent matrix dimensionality, respectively.
Gradient-based optimisation minimises an objective function and the goal is the maximisation of the ELBO, hence the final loss function is defined as the negative of Equation \ref{eq:ELBO}, shown in Equation \ref{eq:TeVAE_loss}.
\begin{equation}
    \mathcal{L}_{\text{VAE}} = -\mathcal{L}_{\theta, \phi}(\textbf{X})
    \label{eq:TeVAE_loss}
\end{equation}

Finally, to enable the backpropagation through the otherwise intractable gradient of the ELBO, the \textit{reparametrisation trick} \cite{kingma_auto-encoding_2014} is applied, shown in Equation \ref{eq:reparametrisation}. This is one of the reasons for the use of a Gaussian distribution as the latent distribution, although any distribution type from the "location-scale" type can be used \cite{kingma_auto-encoding_2014}, like a Laplace distribution.
\begin{equation}
    \textbf{Z} = \boldsymbol{\mu}_\textbf{Z} + \epsilon \cdot \boldsymbol{\sigma}_\textbf{Z}
    \label{eq:reparametrisation}
    \end{equation} 
where $\epsilon \sim \mathcal{N}(0, 1)$ and $(\boldsymbol{\mu}_\textbf{Z}, \log\boldsymbol{\sigma}_\textbf{Z}^2) = q_{\phi}(\textbf{X})$.
\subsection{Multi-head Attention Mechanism}
To simplify the explanation of MA as employed in this work, multi-head self-attention (MS) will be explained instead with the small difference between MA and MS being pointed out at the end.  

MS consists of two different concepts: self-attention and its multi-head extension.
Self-attention is nothing more than scaled dot-product attention \cite{vaswani_attention_2017} where the key, query and value are the same. 
The scaled dot-product attention score is the softmax \cite{bridle_probabilistic_1990} of the product between query matrix $\textbf{Q}$ and key matrix $\textbf{K}$ which is scaled by $\sqrt{d_\textbf{K}}$.
The product between the attention score and the value matrix $\textbf{V}$ yields the context matrix $\textbf{C}$, as shown in Equation \ref{eq:dot_product_attention}.
\begin{equation}
    \textbf{C} = \mathrm{Softmax}\left(\frac{\textbf{Q}\textbf{K}^T}{\sqrt{d_\textbf{K}}}\right)\textbf{V}
    \label{eq:dot_product_attention}
\end{equation}
Compared to recurrent or convolutional layers, self-attention offers a variety of benefits, such as the reduction of computational complexity, as well as an increased amount of operations that can be parallelised \cite{vaswani_attention_2017}.
Also, self-attention inherits an advantage over Bahdanau-style attention \cite{bahdanau_neural_2015} from the underlying scaled dot-product attention mechanism: it can run efficiently in matrix multiplication manner \cite{vaswani_attention_2017}.  

Multi-head self-attention then allows the attention model to attend to different representation subspaces \cite{vaswani_attention_2017}, in addition to learning useful projections rather than it being a stateless transformation \cite{chollet_deep_2021}.
This is achieved using weight matrices $\textbf{W}^Q_i$, $\textbf{W}^K_i$, $\textbf{W}^V_i$, which contain trainable parameters and are unique for each head $i$, as shown in Equation \ref{eq:multi_self_attention}.
\begin{equation}
    \textbf{Q}_i = \textbf{Q}\textbf{W}^Q_i \quad\quad \textbf{K}_i = \textbf{K}\textbf{W}^K_i \quad\quad  \textbf{V}_i = \textbf{V}\textbf{W}^V_i
    \label{eq:multi_self_attention}
\end{equation}
Once the query, key and value matrices are linearly transformed via the weight matrices, the context matrix $\textbf{C}_i$ for each head $i$ is computed using Equation \ref{eq:multi_dot_product_attention}.
\begin{equation}
    \textbf{C}_i = \mathrm{Softmax}\left(\frac{\textbf{Q}_i\textbf{K}_i^T}{\sqrt{d_\textbf{K}}}\right)\textbf{V}_i
    \label{eq:multi_dot_product_attention}
\end{equation}
Then, for $h$ heads, the different context matrices are concatenated and linearly transformed again via the weight matrix $\textbf{W}^O$, resulting in the multi-head context matrix $\textbf{C} \in \mathbb{R}^{w \times d_{\textbf{Z}}}$, Equation \ref{eq:concat_multi_dot_product_attention}.
\begin{equation}
    \textbf{C} = [\textbf{C}_1, ..., \textbf{C}_h]\textbf{W}^{O}
    \label{eq:concat_multi_dot_product_attention}
\end{equation}
The underlying mechanism of MA is identical to MS, with the only difference being that \textbf{K} $=$ \textbf{Q} $\not =$ \textbf{V}. 
\lc{Essentially, MA finds which time steps correlate most with each other inside a given input window and weighs the time steps in context matrix $\textbf{C}$ accordingly.} 
The benefit of this alteration is discussed in Section \ref{sec:proposed_approach}.

\section{Related Work}\label{sec:related_work}
TeVAE belongs to the so-called generative model class, which encompasses both variational autoencoders, as well as generative adversarial networks. 
This section focuses solely on the work on VAE proposed in the context of time-series anomaly detection.

In time-series anomaly detection literature, the only other model that uses the combination of a VAE and an attention mechanism is by \cite{pereira_unsupervised_2018}.
For the purpose of our paper, \lc{it is referred to as variational self-attention VAE (VS-VAE)}.
Their approach consists of a BiLSTM encoder and decoder, where, for an input window of length $w$, the $t=w$ encoder hidden states of each direction are passed on to the variational self-attention (VS) mechanism \cite{bahuleyan_variational_2018}.
The resulting context vector is then concatenated with the sampled latent vector and then passed on to the decoder.
The author claims that applying VS to the VAE model solves the bypass phenomenon, however, no evidence for this claim is provided.

The first published time-series anomaly detection approach based on VAE is LSTM-VAE \cite{park_multimodal_2018}.
One of the contributions is its use of a dynamic prior, i.e.\ $\mathcal{N}(\mu_p, 1)$, rather than a static one, i.e.\ $\mathcal{N}(0, 1)$.
In addition to that, they introduce a state-based threshold estimation method consisting of a support-vector regressor (SVR), which maps the latent distribution parameters $(\mu_{\textbf{z}}, \sigma_{\textbf{z}})$ to the resulting anomaly score using the validation data.
Hence, the dynamic threshold can be obtained through Equation \ref{eq:svr}.
\begin{equation}
    \eta_t = \text{SVR}(\mu_{\textbf{z}, t}, \sigma_{\textbf{z}, t}) + c
    \label{eq:svr}
\end{equation}
where $c$ is a pre-defined constant to control sensitivity.

OmniAnomaly \cite{su_robust_2019} attempts to create a temporal connection between latent distributions by applying a linear Gaussian state space model to them.
For the purpose of this paper, it is abbreviated to OmniA.
Also, it concatenates the last gated recurrent unit (GRU) hidden state with the latent vector sampled in the previous time step.
In addition to that, it uses planar normalising flow \cite{rezende_variational_2015} by applying $K$ transformations to the latent vector in order to approximate a non-Gaussian posterior, as shown in Equation \ref{eq:pnf}.
\begin{equation}
    f^k(\textbf{z}_t^{k-1}) = \textbf{u}\tanh(\textbf{w}\textbf{z}^{k-1}_t)+\textbf{b}
    \label{eq:pnf}
\end{equation}
where $\textbf{u}$, $\textbf{w}$ and $\textbf{b}$ are trainable parameters.

A simplified VAE architecture \cite{pereira_unsupervised_2019} based on BiLSTM layers is also proposed.
For the purpose of our paper, it is called Wasserstein VAE (W-VAE). Unlike its predecessor \cite{pereira_unsupervised_2018}, it drops the attention mechanism but provides contributions elsewhere.
It offers two strategies to detect anomalies based on the VAE outputs.
The first involves clustering the space characterised by the mean parameter of the latent distribution into two clusters and labelling the larger one as normal.
This strategy has a few weaknesses: it cannot be used in an operative environment as it requires some sort of history of test windows to form the clusters and it assumes that there are always anomalous samples present.
The second strategy finds the Wasserstein similarity measure between the latent mean space mapping of the test window in question and the respective mapping $i$ resulting from a representative data subset, such as the validation subset.
Equation \ref{eq:wasserstein} shows how the Wasserstein similarity measure is computed
\begin{equation}
    W_i(\textbf{z}_{\text{test}}, \textbf{z}_{i}) = \|\mu_{\textbf{z}_\text{test}} - \mu_{\textbf{z}_{i}}\|_2^2 + \|\Sigma_{\textbf{z}_\text{test}}^{1/2} - \Sigma_{\textbf{z}_{i}}^{1/2}\|_F^2
    \label{eq:wasserstein}
\end{equation}
where the first term represents the $L2$-Norm between the mean distribution parameters resulting from the test window and each point of the representative subset.
The second term represents the Frobenius norm between the covariance matrix resulting from the test window and each point of the representative subset.

Sliding-window convolutional variational autoencoder (SWCVAE) \cite{chen_unsupervised_2020} is the first that applies convolutional neural networks (CNN) to VAEs for multivariate time-series anomaly detection.
Peculiarly, 2D CNN layers are used with the justification of being able to process the input both spatially and temporally.
We, however, doubt the ability of the model to properly detect anomalies through spatial processing, as a kernel moving along the feature axis can only capture features adjacent to each other.
To create a continuous anomaly score from windows they append the last value of each window to the previous one.
For the purpose of this paper, this process is referred to as \textit{last-type reverse-windowing}.

Smoothness-inducing sequential VAE (SISVAE) \cite{li_anomaly_2021} tries to improve the modelling robustness by the addition of a smoothing term in the loss function which contributes to the reduction of sudden changes in the reconstructed signal, making it less sensitive to noisy time steps.

As part of the variational autoencoder-based selective prediction (VASP) framework \cite{von_schleinitz_vasp_2021}, a variational autoencoder architecture is proposed to increase the robustness of time-series prediction when faced with anomalies.
While the main contribution is attributed to the framework itself, not the VAE, it should be noted that during inference only the mean parameter of the latent distribution is passed to the decoder.

\lc{FedAnomaly \cite{zhang_federated_2021} first extends time-series anomaly detection to a federated learning setting, which may be relevant in use cases involving large amounts of data or strict privacy settings.
Other than being the first VAE based on convolutional GRU layers, the contributions of this paper lie exclusively within the federated framework.
Results show that as a single entity, it performs better than in a federated scenario.}

\lc{Other than a feature selection procedure based on the Kolmogorow-Smirnow test, the lightweight LSTM-VAE (LW-VAE) \cite{fahrmann_lightweight_2022} offers no significant contributions besides being relatively parameter-light.}

\section{Proposed Approach}\label{sec:proposed_approach}
\subsection{Overview}
\begin{figure}[b!]
    \centering
    \def\svgwidth{\textwidth}
    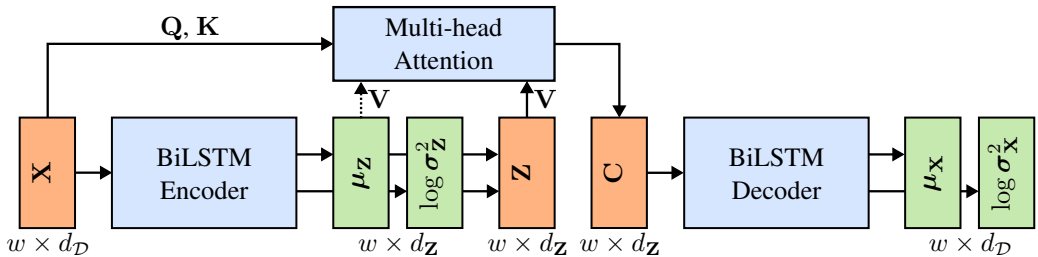
    \caption{An illustration of the proposed TeVAE model. Blue shapes designate trainable models, orange deterministic tensors and green distribution parameters. The shape of each tensor is designated below it. During training $\textbf{Z}$ is used as the value matrix, denoted by the solid arrow, whereas during inference $\boldsymbol{\mu}_\textbf{Z}$ is used as the value matrix, denoted by the traced arrow. Topologically, TeVAE resembles MA-VAE \cite{correia_ma-vae_2023}.} 
    \label{fig:TeVAE_diagram_training}
\end{figure}
\noindent To detect anomalies in multivariate time-series data, we propose a variational autoencoder architecture consisting of BiLSTM layers.
The model architecture is illustrated in Figure \ref{fig:TeVAE_diagram_training}.
During training, the encoder $q_{\phi}$\lc{, parameterised by $\phi$,} maps input window $\textbf{X}$ to a temporal distribution with parameters $\boldsymbol{\mu}_\textbf{Z}$ and $\log\boldsymbol{\sigma}_\textbf{Z}^2$ in the forward pass, Equation \ref{eq:enc_fwd1}.
\begin{equation}
    (\boldsymbol{\mu}_\textbf{Z}, \log\boldsymbol{\sigma}_\textbf{Z}^2) = q_{\phi}(\textbf{X})
    \label{eq:enc_fwd1}
\end{equation}
Given the latent distribution parameters $\boldsymbol{\mu}_\textbf{Z}$ and $\log\boldsymbol{\sigma}_\textbf{Z}^2$, the latent matrix is sampled from the resulting distribution, as shown in Equation \ref{eq:enc_fwd2}. \lc{Note that the covariance is not modelled, hence $\boldsymbol{\sigma}_\textbf{Z}$ only contains the diagonal of the covariance matrix.}
\begin{equation}
    \textbf{Z} \sim \mathcal{N}(\boldsymbol{\mu}_\textbf{Z}, \text{diag}(\boldsymbol{\sigma}_\textbf{Z}))
    \label{eq:enc_fwd2}
\end{equation}
Then, the input window $\textbf{X}$ is linearly transformed to obtain the query matrices $\textbf{Q}_i$ and key matrices $\textbf{K}_i$ for each head $i$.
Likewise, the sampled latent matrix $\textbf{Z}$ is also transformed to the value matrix $\textbf{V}_i$, as shown in Equation \ref{eq:att_fwd1}.
\begin{equation}
    \textbf{Q}_i = \textbf{X}\textbf{W}^Q_i \quad\quad \textbf{K}_i = \textbf{X}\textbf{W}^K_i \quad\quad \textbf{V}_i = \textbf{Z}\textbf{W}^V_i
    \label{eq:att_fwd1}
\end{equation}
To output the context matrix $\textbf{C}_i$ for each head $i$, the softmax of the through $\sqrt{d_\textbf{K}}$ normalised query and key product is multiplied with the value matrix, Equation \ref{eq:att_fwd2}.
\begin{equation}
    \textbf{C}_i = \mathrm{Softmax}\left(\frac{\textbf{Q}_i\textbf{K}_i^T}{\sqrt{d_\textbf{K}}}\right)\textbf{V}_i
    \label{eq:att_fwd2}
\end{equation}
The final context matrix $\textbf{C}$ is the result of the linearly-transformed concatenation of each head-specific context matrix $\textbf{C}_i$, as expressed in Equation \ref{eq:att_fwd3}.
\begin{equation}
    \textbf{C} = [\textbf{C}_1, ..., \textbf{C}_h]\textbf{W}^{O}
    \label{eq:att_fwd3}
\end{equation}
The decoder $p_{\theta}$\lc{, parameterised by $\theta$,} then maps the context matrix $\textbf{C}$ to $\boldsymbol{\mu}_\textbf{X}$ and $\log\boldsymbol{\sigma}_\textbf{X}^2$, as shown in Equation \ref{eq:dec_fwd1}. These can then be used to parametrise output distribution \lc{$\mathcal{N}(\boldsymbol{\mu}_\textbf{X}, \text{diag}(\boldsymbol{\sigma}_\textbf{X}))$}.
\begin{equation}
    (\boldsymbol{\mu}_\textbf{X}, \log\boldsymbol{\sigma}_\textbf{X}^2) = p_{\theta}(\textbf{C})
    \label{eq:dec_fwd1}
\end{equation}
\lc{Mapping the output to a distribution rather than a deterministic vector allows TeVAE to model uncertainty, which is assumed to be normally distributed.}

\subsection{Inference Mode}
Despite the generative capabilities of VAEs, TeVAE does not leverage generation for anomaly detection.
Rather than sampling a latent matrix as shown in Equation \ref{eq:enc_fwd2} during inference, sampling is disabled and only $\boldsymbol{\mu}_\textbf{Z}$ is taken as the input for the multi-head attention mechanism, like in \cite{von_schleinitz_vasp_2021}.
Equation \ref{eq:enc_fwd2}, therefore, is replaced by Equation \ref{eq:inference} for the forward pass.
\begin{equation}
    \textbf{Z} = \boldsymbol{\mu}_\textbf{Z}
    \label{eq:inference}
\end{equation}
This not only accelerates inference by eliminating the sampling process but is also empirically found to be a good approximation of an averaged latent matrix if it were sampled several times like in \cite{pereira_unsupervised_2018}.
The TeVAE layout during inference is shown in Figure \ref{fig:TeVAE_diagram_training}, where the traced arrow designates the information flow from the encoder to the MA mechanism.

\subsection{Threshold Estimation Method}
Anomalies are by definition very rare events, hence an ideal anomaly detector only flags measurements very rarely but accurately.
In the powertrain test bench scenario an algorithm is preferred that only flags a sequence it is sure is an anomaly, in other words, an algorithm that outputs very few to no false positives.
A high false positive count would lead to a lot of stoppages and therefore lost testing time and additional cost.
Of course, the vast majority of measurements evaluated will be anomaly-free hence it is paramount to classify them correctly, naturally leading to a high precision value.
Also, there is no automatic evaluation methodology currently running at test benches, other than rudimentary rule-based methods, therefore a solution that plugs into the existing system that automatically detects \textit{some} or \textit{most} anomalies undetectable by rules\lc{-based approaches} can already lead to \lc{time and cost savings}.
To achieve this, the threshold $\tau$ is set as the maximum negative log-likelihood observed when the model is fed with \lc{unlabelled} validation data.

\subsection{Bypass Phenomenon}
VAE, when combined with an attention mechanism, can exhibit a behaviour called the bypass phenomenon \cite{bahuleyan_variational_2018}.
When the bypass phenomenon happens the latent path between encoder and decoder is bypassed and information flow occurs mostly or exclusively through the attention mechanism, as it has deterministic access to the encoder hidden states and therefore avoids regularisation through the $D_{\text{KL}}$ term.
In an attempt to avoid this, \cite{bahuleyan_variational_2018} propose variational attention, which, like the VAE, maps the input to a distribution rather than a deterministic vector.
Applied to natural language processing, \cite{bahuleyan_variational_2018} demonstrate that this leads to a diversified generated portfolio of sentences, indicating alleviation of the bypassing phenomenon.
As previously mentioned, only \cite{pereira_unsupervised_2018} applies this insight in the anomaly detection domain, however, they do not present any evidence that it alleviates the bypass phenomenon in their work.
TeVAE on the other hand, cannot suffer from the bypass phenomenon in the sense that information flow ignores the latent variational path between encoder and decoder since the MA mechanism requires the value matrix \textbf{V} from the encoder to output the context matrix. 
Assuming the bypass phenomenon also applies to a case where information flow ignores the attention mechanism, one could claim that TeVAE is not immune. 
To disprove this claim, the attention mechanism is removed from the model in an ablation study to see if anomaly detection performance remains the same. 
In this case, $\textbf{V}=\textbf{Z}$ is instead directly input into the decoder. 
If it drops, it is evidence of the contribution of the attention mechanism to the model performance and hence is not bypassed. 
The results for this ablation study are shown and discussed in Section \ref{sec:results}.

\subsection{Reverse-window Process}
\begin{figure*}[t!]
    \centering
    \def\svgwidth{\textwidth}
    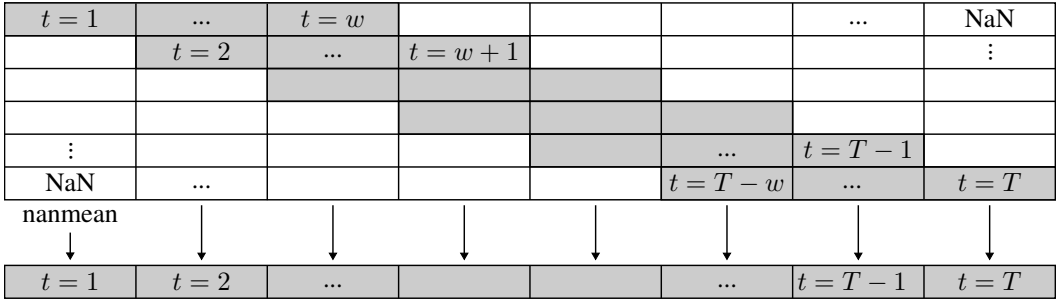
    \caption{\lc{Mean-type reverse windowing process illustrated. The grey boxes represent individual windows.}} 
    \label{fig:nan_mean}
\end{figure*}

Since the model is trained to reconstruct fixed-length windows, the same applies during inference.
However, to decide whether a given measurement sequence $\mathcal{S}_n \in \mathbb{R}^{T_n \times d_\mathcal{D}}$ is anomalous, a continuous reconstruction of the measurement is required.
A trivial way to do so would be to window the input measurement $\mathcal{S}_n$ using a shift of $1$, input the windows into the model and chain the last time step from each output window to obtain a continuous sequence \cite{chen_unsupervised_2020}.
Considering the BiLSTM nature of the encoder and decoder, the first and last time steps of a window can only be computed given the states from one direction, making these values, in theory, less accurate, however.
To overcome this, we propose averaging matching time steps in overlapping windows, which is called \textit{mean-type} reverse-window method.
This is done by pre-allocating an array with NaN values, filling it, and taking the mean for each time step while ignoring the NaN values, as depicted in Figure \ref{fig:nan_mean}.
This process and the general anomaly detection process are described in Algorithm \ref{alg:anomaly_detection}.
\lc{The input for this process is the output distribution parameters $(\boldsymbol{\mu}_{\textbf{X}}, \log \boldsymbol{\sigma}_{\textbf{X}}^2)$, so it essentially averages the distributions and hence can only be applied to the mean and \textit{variance} parameters.
Consider the distributions $\mathcal{N}(\mu_x, \sigma_x^2)$ and $\mathcal{N}(\mu_y, \sigma_y^2)$, both assumed to be independant and normally distributed. The sum of both distributions results in normal distribution $\mathcal{N}(\mu_{x+y}, \sigma_{x+y}^2)$, obtained as shown in Equation \ref{eq:mean_var} \cite{lemons_introduction_2002}.
\begin{equation}
    \mu_{x+y} = \mu_x + \mu_y \quad\quad \sigma_{x+y}^2 = \sigma_x^2 + \sigma_y^2
    \label{eq:mean_var}
\end{equation}
Therefore, the standard deviation of the resulting distribution $\sigma_{x+y}$ is characterised by Equation \ref{eq:std_dev}.
\begin{equation}
    \sigma_{x+y} = \sqrt{\sigma_x^2 + \sigma_y^2}
    \label{eq:std_dev}
\end{equation}
Therefore, to take the mean of the distributions the log variance $\log \boldsymbol{\sigma}_{\textbf{X}}^2$ is converted to the variance $\boldsymbol{\sigma}_{\textbf{X}}^2$, averaged, and then converted to the \textit{standard deviation} $\boldsymbol{\sigma}_{\textbf{X}}$.}

With a continuous mean $\boldsymbol{\mu}_{\mathcal{S}}$ and standard deviation $\boldsymbol{\sigma}_{\mathcal{S}}$, the continuous negative log-likelihood, i.e.\ the anomaly score $\mathbf{s}$, is computed for the respective measurement.
A comparison between the mean, last and first reverse-window process is provided in Section \ref{sec:results}.
\begin{algorithm}[t!]
\caption{Anomaly Detection Process}\label{alg:anomaly_detection}
\begin{algorithmic}
\Input $\text{Sequence  } \mathcal{S}_n \in \mathbb{R}^{T_n \times d_\mathcal{D}},   \text{Threshold  } \tau$ 
\Result $\text{Label   }l_n \text{    for sequence   }\mathcal{S}_n$
\State $n_{\text{windows}} \gets T_n-w+1$ \Comment{\lc{Find the total number of windows in sequence}}

\State $\boldsymbol{\mu}_{\textbf{X}, \text{temp}} \gets \text{zeros}(n_{\text{windows}}, T_n, d_\mathcal{D}) + \text{NaN}$ \Comment{\lc{Pre-allocate NaN array to assign all windows to}}

\State $\boldsymbol{\sigma}_{\textbf{X}, \text{temp}}^{2} \gets \text{zeros}(n_{\text{windows}}, T_n, d_\mathcal{D}) + \text{NaN}$ \Comment{\lc{Pre-allocate NaN array to assign all windows to}}

\For{$i=1 \to n_{\text{windows}}$} \Comment{\lc{Iterate through total number of windows}}
    \State $\textbf{X} \gets \mathcal{S}[i:w+i]$ \; \Comment{\lc{Assign current window}}
    
    \State $(\boldsymbol{\mu}_{\textbf{Z}}, \log \boldsymbol{\sigma}_{\textbf{Z}}^2) \gets q_{\phi}(\textbf{X})$\; \Comment{\lc{Input window into encoder}}
    
    \State $\textbf{C} \gets \text{MA}(\textbf{X}, \textbf{X}, \boldsymbol{\mu}_{\textbf{Z}})$\; \Comment{\lc{Input window and mean parameter into MA}}
    
    \State $(\boldsymbol{\mu}_{\textbf{X}}, \log \boldsymbol{\sigma}_{\textbf{X}}^2) \gets p_{\theta}(\textbf{C})$\; \Comment{\lc{Input context matrix into decoder}}

    \State $\boldsymbol{\mu}_{\textbf{X}, \text{temp}}[i, i: i+w] \gets \boldsymbol{\mu}_{\textbf{X}}$ \Comment{\lc{Assign mean parameter to pre-allocated array}}
    
    \State $\boldsymbol{\sigma}_{\textbf{X}, \text{temp}}^{2}[i, i: i+w] \gets \boldsymbol{\sigma}_{\textbf{X}}^2$ \Comment{\lc{Assign variance parameter to pre-allocated array}}

\EndFor

\State $\boldsymbol{\mu}_{\mathcal{S}} \gets \text{nanmean}(\boldsymbol{\mu}_{\textbf{X}, \text{temp}}, 0)$ \Comment{\lc{Take nanmean along axis 0}}

\State $\boldsymbol{\sigma}_{\mathcal{S}}^2 \gets \text{nanmean}(\boldsymbol{\sigma}_{\textbf{X}, \text{temp}}^2, 0)$ \Comment{\lc{Take nanmean along axis 0}}
    
\State $\mathbf{s}_n \gets -\log p(\boldsymbol{\mu}_{\mathcal{S}}, \boldsymbol{\sigma}_{\mathcal{S}}\vert \mathcal{S}_n )$\; \Comment{\lc{Obtain continuous negative log-likelihood (anomaly score)}}

\State $l_n \gets \text{max}(\mathbf{s}_n) > \tau$\; \Comment{\lc{Compare maximum value in anomaly score with threshold}}
\end{algorithmic}
\end{algorithm}

\begin{figure*}[t!]
    \centering
    \def\svgwidth{\textwidth}
    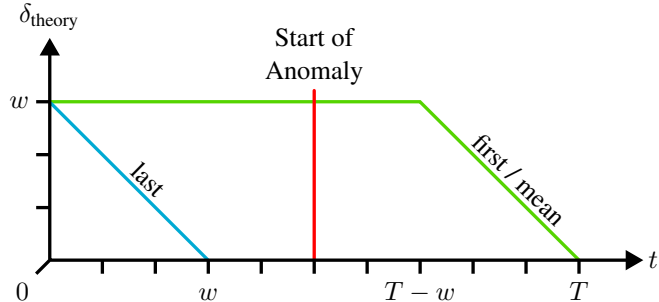
    \caption{\lc{Theoretical delay $\delta_\text{theory}$ plotted against time $t$. The last-type reverse-windowing is represented by a solid blue line and the first and mean-type by a solid green line. The red line represents the start of a hypothetical sub-sequence anomaly.}}
    \label{fig:delay2}
\end{figure*}

\lc{The theoretical delay $\delta_\text{theory}$ associated with each of the reverse-window processes can be discussed ahead of Section \ref{sec:results}, however.
$\delta_\text{theory}$ is defined as the intrinsic delay introduced by each reverse-windowing method.
To illustrate the theoretical delay $\delta_\text{theory}$, it is plotted against time $t$ to demonstrate the delay for each of the reverse-window processes, shown in Figure \ref{fig:delay2}.
For the last-type reverse-window method during $0<t<w$, no time steps can be evaluated until a full window can be formed, streamed and evaluated, introducing a theoretical delay $\delta_\text{theory}=w$. 
This property is intrinsic to approaches based on fixed-length windows rather than variable-length sequences. 
At time step $t=w$, however, all time steps $0<t<w$ are evaluated and output at the same time.
For $w<t<T$, however, the last time step of each window corresponds the current real-world time step, i.e.\ $\delta_\text{theory}=0$ for $w<t<T$.
As mentioned above, the lack of evaluation until $t=w$ is natural to any window-based approach and hence the first-type and mean-type reverse-window methods show the same behaviour.
In contrast to last-type reverse-windowing, however, these methods only output the first value of each window i.e.\ evaluation is $\delta_\text{theory}=w$ time steps behind for $0<t<T-w$.
At $t=T$, though, the time steps $T-w<t<T$ are all output at the same time, meaning that $\delta_\text{theory}=0$ and no extra time is needed for evaluation after streaming ends.
It should be noted, however, that in online time-series anomaly detection the last-type reverse-window method is in theory faster than the other two types.
Consider a sub-sequence anomaly starting at time step $w+2$, designated by the red line in Figure \ref{fig:delay2}.
Apart from the time required for inference, the last-type reverse-window method can detect the anomaly without a theoretical delay $\delta_\text{theory}$, during $w<t<T$, whereas the other two methods can only detect the anomaly $\delta_\text{theory}=w$ time steps later. 
For $0<t<T-w$, a theoretical delay of $\delta_\text{theory}=w$ time steps is, therefore, the absolute best that first-type and mean-type reverse-windowing can achieve, however, while the best delay the first-type can achieve is $\delta_\text{theory}=0$, it will be higher in reality, perhaps higher than $w$ time steps.}

\subsection{Root-cause Analysis}
\lc{Once a measurement is predicted to be anomalous it is of great benefit if context is provided, like what channel had the biggest impact on the prediction.
Depending on what subsystem within the system the anomaly originates, it may affect a different number of channels, in some cases even all.
Some attempts to provide root-cause functionality are made in literature.} 

\lc{The first known to the authors is proposed by \cite{zhang_deep_2019}.
The approach does not use time series but instead, the resulting correlation matrices and the anomaly score is given as the difference between the input and reconstructed correlation matrices.
Hence, as a mean of providing root-cause information they rank each channel by the number of high anomaly scores for every other channel.
Root-cause identification is then quantified by the recall for the top $k$ channels, in this case three, although little information on how exactly this is calculated is provided.
Another issue with this metric is that is not parameter-free, since $k$ needs to be set manually.
Furthermore, the method used is specific to approaches using the same type of correlation matrices, which is rare in literature.}

\lc{Approaches using the negative log-likelihood as the anomaly score obtain it from a multivariate output distribution with a diagonal covariance.
While the resulting anomaly score is a univariate time series, it is referred to as \textit{multivariate} anomaly score for now due to its origin in a multivariate output distribution.
\cite{su_robust_2019} propose decomposing the anomaly score into anomaly scores for each channel, by splitting the multivariate output distribution into $d_\mathcal{D}$ univariate output distributions.
The resulting anomaly scores for each of the $d_\mathcal{D}$ channels are referred to as \textit{univariate} anomaly scores due to their origin in $d_\mathcal{D}$ univariate output distributions.
To find the channel that contributed most to a detection, the highest univariate anomaly score is assumed to be the root cause.}
 
\lc{In this work, the methodology proposed by \cite{su_robust_2019} is used to find the channels that contribute most to a detection. 
To this end, the method is slightly adapted, given the online premise of the use case.
Rather than considering the univariate anomaly scores as time series for root cause analysis, the first time step of the multivariate anomaly score above the threshold is used.
Then, for that time step, the highest univariate anomaly score is assumed as the contributing channel.}

\section{Results}\label{sec:results}
\subsection{Setup}
The encoder and decoder both consist of two BiLSTM layers, with the outer ones having 512 hidden- and cell-state sizes and the inner ones 256.
All other parameters are left as the default in the TensorFlow API. 

During training only, input windows are corrupted using Gaussian noise using $0.01$ standard deviation to increase robustness \cite{vincent_extracting_2008}.

Key factors that are investigated in Section \ref{sec:results} are given a default value which applies to all experiments unless otherwise specified.
These factors are training and validation subset size, which is set to $512$h, reverse-window method, where the mean-type is used, the latent dimension size, which is set to $d_{\textbf{Z}}=64$, the MA mechanism, which is set up as proposed in \cite{vaswani_attention_2017} with a head count of $h=8$ and a key dimension size $d_\textbf{K}=\lfloor d_\mathcal{D}/h \rfloor=1$.

The optimiser used is the AMSGrad optimiser with the default parameters in the TensorFlow API.

Cyclical $D_{\text{KL}}$ annealing is applied to the training of TeVAE, to avoid the $D_{\text{KL}}$ vanishing problem \cite{fu_cyclical_2019}.
The $D_{\text{KL}}$ vanishing problem occurs when regularisation is too strong at the beginning of training, i.e.\ the Kullback-Leibler divergence term has a larger magnitude in relation to the reconstruction term.
Cyclical $D_{\text{KL}}$ annealing allows the model to weigh the Kullback-Leibler divergence lower than the reconstruction term in a cyclical manner through a weight $\beta$.
This callback is configured with a grace period of 25 epochs, where $\beta$ is linearly increased from $0$ to $10^{-8}$.
After the grace period, $\beta$ is set to $10^{-8}$ and is gradually increased linearly to $10^{-2}$ throughout the following $25$ epochs, representing one loss cycle.
This loss cycle is repeated until the training stops.

All priors in this work are set as standard Gaussian distributions, i.e.\ $p = \mathcal{N}(0,1)$.

To prevent overfitting, early stopping is implemented.
It works by monitoring the negative log-likelihood component of the validation loss during training and stopping if it does not improve for $250$ epochs.
Logically, the model weights at the lowest validation negative log-likelihood are saved.

\lc{Given the stochastic nature of the VAEs, the chosen seed can impact the anomaly detection performance as it can lead to a different local minimum during training, hence all tests are done with three different seeds and are shown in form of the standard deviation after every performance metric.}

Training is done on a workstation configured with an NVIDIA RTX A6000 GPU.
The library used for model training is TensorFlow 2.10.1 on Python 3.10 on Windows 10 Enterprise LTSC version 21H2.

\subsection{Online Evaluation Metrics}
The results provided are given in the form of the calibrated and uncalibrated anomaly detection performance, i.e.\ with and without consideration of threshold $\tau$, respectively.
Recall that the threshold used is the maximum negative log-likelihood obtained from the validation set. \lc{The basis for all metrics are the number of true positives $N_\text{tp}$, number of false negatives $N_\text{fn}$ and number of false positives $N_\text{fp}$.}

\lc{As discussed in Section \ref{sec:introduction}, a testing subset in discrete time-series anomaly detection problem has three types of time series: entirely normal, time-series anomalies and sub-sequence anomalies.
Since anomalies are considered rare events, the number of anomaly-free time series $N_{\text{af}}$ within $\mathcal{D}^{\text{test}}$ is much larger than the number of time-series anomalies $N_{\text{ts}}$ and sub-sequence anomalies $N_{\text{ss}}$, such that $N_{\text{af}}>>N_{\text{ts}}+N_{\text{ss}}=N_{\text{a}}$ and $N=N_{\text{af}}+N_{\text{ts}}+N_{\text{ss}}$.
In the case of anomaly-free time series and time-series anomalies, traditional labels can easily be applied. 
An anomaly-free time series can be labelled as true negative or false positive and a time-series anomaly can be labelled as true positive or false negative. 
For partially anomalous time series, i.e.\ where the anomalous behaviour occupies a contiguous subset of time steps within the time series, it can be labelled as a true positive or a false negative, but also as a false positive, which occurs when an algorithm flags a time step early.
Formally, this is the case when the first flagged time step is far enough ahead of the first ground-truth time step that the model cannot have had access to it.
As is evident, the proposed metrics can only be applied to discrete time-series data sets where anomalous time series have at most one contiguous ground-truth anomalous sub-sequence of any length. 
Also, ensuring each time series contains at most one contiguous anomaly avoids ambiguity on how multiple sub-sequence anomalies within a time series should be detected and counted \cite{wu_current_2021}.
In addition to that, there can also be at most one contiguous ground-truth anomaly-free sub-sequence of any length within a sub-sequence anomalous time series, since systems that can dynamically return to anomaly-free behaviour (especially in a short amount of time) reap less benefit from automated anomaly detection than those which require human attention. 
Given that a predicted anomaly likely requests human attention and potentially a stoppage of the system, the adaptation of the metrics proposed above only takes into account the first predicted anomalous time step. 
For cases where the process may continue, there can be multiple contiguous predicted anomalous sub-sequences, hence the metrics cannot be applied.}

Calibrated metrics are the precision, recall and $F_1$ score.
Precision $P$ represents the ratio between \lc{the number of} correctly identified anomalies (true positives) and \lc{the number of} all positives (true and false), shown in Equation \ref{eq:precision_recall_f1}, recall $R$ represents the ratio between \lc{the number of} true positives and \lc{the number of} all anomalies, shown in Equation \ref{eq:precision_recall_f1}, and $F_1$ score represents the harmonic mean of the precision and recall, shown in Equation \ref{eq:precision_recall_f1}.
\begin{equation}
    P = \frac{N_\text{tp}}{N_\text{tp}+N_\text{fp}} \quad\quad R = \frac{N_\text{tp}}{N_\text{tp}+N_\text{fn}} \quad\quad F_1 = 2\cdot\frac{P\cdot R}{P+R}
    \label{eq:precision_recall_f1}
\end{equation}

The theoretical maximum $F_1$ score, $F_{1,\text{best}}$, is also provided to aid discussion.
This represents the best possible score achievable by the approach if the ideal threshold were known, i.e.\ the point on the precision-recall curve that comes closest to the $P=R=1$ point, though, in reality, this value is not observable and hence \textit{cannot} be obtained in an unsupervised manner. 
\lc{The precision and recall corresponding to the $F_{1,\text{best}}$ score are also provided.}

The uncalibrated anomaly detection performance, i.e.\ the performance for a range of thresholds is represented by the area under the continuous precision-recall curve $A_\text{PR}^\text{cont}$, Equation \ref{eq:cont_aucpr}.
\begin{equation}
    A_\text{PR}^\text{cont}=\int_0^1 P \,dR 
    \label{eq:cont_aucpr}
\end{equation}
As the integral cannot be computed for the continuous function, the area under the discrete precision-recall curve $A_\text{PR}^\text{disc}$ is used which is done using the trapezoidal rule, Equation \ref{eq:disc_aucpr}.
\begin{equation}
    A_\text{PR}^\text{disc}=\sum^{K-1}_{k=1}\frac{P_{k-1}+P_{k}}{2}\Delta R_k
    \label{eq:disc_aucpr}
\end{equation}
where $K$ is the number of discrete points along the precision-recall curve, $k$ the index of discrete points along the precision-recall curve and $\Delta R_k$ the sub-interval length between indices $k$ and $k-1$.

\lc{While the above metrics quantify binary anomaly detection performance, they do not provide information on the delay of detections, which plays a crucial role in online time-series anomaly detection. 
Therefore, we propose an additional metric, the detection delay $\delta$, which represents the absolute delay between the first ground-truth anomalous time step $t=t_{\text{gt}}$ and the first predicted anomalous time step $t=t_{\text{p}}$. 
The detection delay $\delta$ is only calculated for anomalous time series since in anomaly-free time series there is no first ground-truth time step $t=t_{\text{gt}}$, therefore it is calculated as shown in Equation \ref{eq:detection_delay}.
\begin{equation}
         \delta = \vert t_{\text{p}} - t_{\text{gt}} \vert
    \label{eq:detection_delay}
\end{equation}
where, in case of a false negative, $t_{\text{p}}$ is equal to the last time step of the anomalous time series. 
This formula not only reflects the detection delay for true positives but also punishes false negatives by applying the maximum delay possible, i.e.\ $\delta=T$, as well as false positives by applying the "negative" delay between the first early predicted time step $t=t_{\text{p}}$ and the first ground-truth time step $t=t_{\text{gt}}$.
For each anomalous test time series the detection delay is calculated and subsequently averaged to yield the average detection delay $\bar{\delta}$, shown in equation \ref{eq:average_detection_delay}. 
\begin{equation}
    \bar{\delta} = \frac{1}{N_{\text{ts}}+N_{\text{ss}}}\sum^{N_{\text{ts}}+N_{\text{ss}}}_{i=1} \delta_i
    \label{eq:average_detection_delay}
\end{equation}}

\lc{In order to evaluate TeVAE's root-cause analysis performance we propose a corresponding metric. 
Recall that according to the method of counting labels, anomaly-free sequences can be labelled as true negatives or false positives and time-series anomalies can be labelled as true positives or false negatives. 
Sub-sequence anomalies can be labelled as true positives, false negatives or false positives, where the latter occurs when an anomaly is detected before it actually occurs.
To quantify whether the most relevant channel has been flagged in the case of a predicted anomalous sequence, the root-cause true positive count $N_{\text{tp}_\text{rc}}$ and root-cause false positive count $N_{\text{fp}_\text{rc}}$ are introduced.
Clearly, the root-cause channel can only be obtained for predicted anomalous sequences, hence why there is no root-cause true negative count $\text{tn}_\text{rc}$ or root-cause false negative count $N_{\text{fn}_\text{rc}}$.
The root-cause true positive count $N_{\text{tp}_\text{rc}}$ represents the number of sequences labelled as true positives and the predicted root-cause channel is a subset of the list of ground-truth root-cause channels for a given anomaly type.
Likewise, root-cause false positive count $N_{\text{fp}_\text{rc}}$ represents the sum of three cases. 
The first case is an anomaly-free sequence labelled as a false positive, for which there are no ground-truth root-cause channels.
The second case is a ground-truth time-series or sub-sequence anomaly that is labelled as a true positive, but the predicted root-cause channel is not a subset of the list of ground-truth root-cause channels for the relevant anomaly type.
The third case is a ground-truth sub-sequence anomaly that is labelled as a false positive, due to a premature detection, for which there are no ground-truth root-cause channels.
To aid the understanding of this concept, a diagram depicting what types of sequences can be labelled as what is shown in Table \ref{tab:root_cause}.}
\begin{table}[h!]
\centering
\caption{\lc{Table depicting which sequence types can be classified with what detection labels and root-cause labels.}}
\label{tab:root_cause}
\begin{tabular}{lcccc}
\begin{tabular}[c]{@{}c@{}}Sequence  Type\end{tabular}       &             & \begin{tabular}[c]{@{}c@{}}Detection Label\end{tabular} &             & \begin{tabular}[c]{@{}c@{}}Root-cause Label\end{tabular} \\ \hline\hline
\begin{tabular}[c]{@{}c@{}}Normal sequence\end{tabular}      & $\rightarrow$ & \begin{tabular}[c]{@{}c@{}}true negative\end{tabular}   & $\rightarrow$ & -                                                          \\
                                                               & \rotatebox[origin=c]{180}{$\Lsh$}	    & \begin{tabular}[c]{@{}c@{}}false positive\end{tabular}  & $\rightarrow$ & \begin{tabular}[c]{@{}c@{}}false positive\end{tabular}   \\\hline
\begin{tabular}[c]{@{}c@{}}Time-series anomaly\end{tabular}  & $\rightarrow$ & \begin{tabular}[c]{@{}c@{}}false negative\end{tabular}  & $\rightarrow$ & -                                                          \\
                                                               & \rotatebox[origin=c]{180}{$\Lsh$}	    & \begin{tabular}[c]{@{}c@{}}true positive\end{tabular}   & $\rightarrow$ & \begin{tabular}[c]{@{}c@{}}true positive\end{tabular}    \\
                                                               &             &                                                           & \rotatebox[origin=c]{180}{$\Lsh$}	    & \begin{tabular}[c]{@{}c@{}}false positive\end{tabular}   \\\hline
\begin{tabular}[c]{@{}c@{}}Sub-sequence anomaly\end{tabular} & $\rightarrow$ & \begin{tabular}[c]{@{}c@{}}false negative\end{tabular}  & $\rightarrow$ & -                                                          \\
                                                               & \rotatebox[origin=c]{180}{$\Lsh$}	    & \begin{tabular}[c]{@{}c@{}}false positive\end{tabular}  & $\rightarrow$ & \begin{tabular}[c]{@{}c@{}}false positive\end{tabular}   \\
                                                               & \rotatebox[origin=c]{180}{$\Lsh$}	    & \begin{tabular}[c]{@{}c@{}}true positive\end{tabular}   & $\rightarrow$ & \begin{tabular}[c]{@{}c@{}}true positive\end{tabular}    \\
                                                               &             &                                                           & \rotatebox[origin=c]{180}{$\Lsh$}	    & \begin{tabular}[c]{@{}c@{}}false positive\end{tabular}  \\\hline
\end{tabular}
\end{table}
\lc{As is evident, $N_\text{tp} + N_\text{fp} = N_{\text{tp}_\text{rc}} + N_{\text{fp}_\text{rc}}$.
To summarise the $N_{\text{tp}_\text{rc}}$ and $N_{\text{fp}_\text{rc}}$ figures, we propose a new metric called the root-cause precision $P_\text{rc}$, shown in Equation \ref{eq:root-cause}.
\begin{equation}
    P_\text{rc} = \frac{N_{\text{tp}_\text{rc}}}{N_\text{tp}+N_\text{fp}}
    \label{eq:root-cause}
\end{equation}
$P_\text{rc}$ denotes the number of correctly identified root-cause channels relative to the number of total detections, true or false. 
}

\subsection{Ablation Study}
TeVAE is tested without the MA mechanism and with a direct connection from the encoder to the decoder to observe whether the absence of the MA impacts results. The anomaly detection performance of TeVAE and its counterpart without MA, henceforth referred to as \textit{NoMA} model, are shown in Table \ref{tab:ablation}.
\begin{table}[h!]
\centering
\caption{\lc{$F_1$ score, precision $P$, recall $R$, average detection delay $\bar{\delta}$ and root-cause precision $P_\text{rc}$ using the unsupervised threshold (top half) and theoretical best threshold (bottom half), as well as the area under the precision-recall curve $A_\text{PR}$ for NoMA and TeVAE. The best values for each metric are given in \textbf{bold}. The standard deviation for the different seeds are also provided.}}\label{tab:ablation}
\setlength{\tabcolsep}{3pt}
\begin{tabular}{lcccccc}
Model   & $F_1$                         & $P$                       & $R$                      & $A_{\text{PR}}$           & $\bar{\delta} \ [s]$      & $P_\text{rc}$             \\ \hline\hline
NoMA    & $0.59 \pm 0.01$               & $\textbf{0.98} \pm 0.02$  & $0.42 \pm 0.01$          & $0.56 \pm 0.01$           & $577.0 \pm \phantom 05.8$ & $\textbf{0.85} \pm 0.04$  \\
TeVAE   & $\textbf{0.70} \pm 0.03$      & $0.92 \pm 0.08$           & $\textbf{0.57} \pm 0.04$ & $\textbf{0.66} \pm 0.04$  & $\textbf{418.1} \pm 17.3$ & $0.63 \pm 0.18$           \\ \hline
          
NoMA    & $0.59 \pm 0.01$               & $\textbf{1.00} \pm 0.00$  & $0.42 \pm 0.01$          & $0.56 \pm 0.01$            & $577.2 \pm \phantom 05.7$  & $\textbf{0.86} \pm 0.03$\\
TeVAE   & $\textbf{0.72} \pm 0.03$      & $0.97 \pm 0.04$           & $\textbf{0.58} \pm 0.06$ & $\textbf{0.66} \pm 0.04$   & $\textbf{412.9} \pm 25.0$  & $0.67 \pm 0.15$         \\\hline
\end{tabular}
\end{table}

While the precision value of the NoMA model is slightly higher than the TeVAE, the recall value on the other hand is much lower.
Overall, TeVAE has a significantly higher $F_1$ score, as well as a higher theoretical maximum $F_1$ score \lc{and higher uncalibrated anomaly detection performance, denoted by the $A_{\text{PR}}$ figure.
Furthermore, TeVAE features a much lower average detection delay which is especially relevant for online time-series anomaly detection.
In contrast to that, NoMA offers marginally higher root-cause precision.}
The results hence point towards an improvement brought about by the addition of the MA mechanism and therefore the bypass phenomenon can be ruled out.

\subsection{Data Set Size Requirements}
To evaluate how much data is required to train TeVAE to a point of adequate anomaly detection performance, it has been trained with $1$h, $8$h, $64$h, and $512$h of dynamic testing time. The results for this experiment are presented in Table \ref{tab:dataset}.
\begin{table}[h!]
\centering
\caption{\lc{$F_1$ score, precision $P$, recall $R$, average detection delay $\bar{\delta}$ and root-cause precision $P_\text{rc}$ using the unsupervised threshold (top half) and theoretical best threshold (bottom half), as well as the area under the precision-recall curve $A_\text{PR}$ for the different training and validation subset sizes. The best values for each metric are given in \textbf{bold}. The standard deviation for the different seeds are also provided.}}\label{tab:dataset} 
\setlength{\tabcolsep}{3pt}
\begin{tabular}{lcccccc}
Size      & $F_1$                     & $P$                         & $R$                       & $A_{\text{PR}}$          & $\bar{\delta} \ [s]$          & $P_\text{rc}$    \\ \hline\hline
$1$h      & $0.20 \pm 0.02$           & $0.11 \pm 0.01$             & $\textbf{0.83} \pm 0.06$  & $0.57 \pm 0.02$           & $321.1 \pm 30.4$              & $0.07 \pm 0.01$   \\
$8$h      & $0.37 \pm 0.14$           & $0.26 \pm 0.13$             & $0.80 \pm 0.06$           & $0.69 \pm 0.04$           & $\textbf{298.7} \pm 29.2$     & $0.16 \pm 0.08$   \\
$64$h     & $0.69 \pm 0.06$           & $0.76 \pm 0.15$             & $0.66 \pm 0.03$           & $\textbf{0.71} \pm 0.01$  & $371.3 \pm 38.4$              & $0.61 \pm 0.15$   \\
$512$h    & $\textbf{0.70} \pm 0.03$  & $\textbf{0.92} \pm 0.08$    & $0.57 \pm 0.04$           & $0.66 \pm 0.04$           & $418.1 \pm 17.3$              & $\textbf{0.63} \pm 0.18$ \\ \hline

$1$h      & $0.61 \pm 0.05$           & $0.72 \pm 0.04$             & $0.55 \pm 0.09$           & $0.57 \pm 0.02$           & $459.9 \pm 42.5$              & $0.54 \pm 0.07$   \\
$8$h      & $0.74 \pm 0.04$           & $0.94 \pm 0.02$             & $0.61 \pm 0.06$           & $0.69 \pm 0.04$           & $460.9 \pm 35.1$              & $0.68 \pm 0.04$    \\
$64$h     & $\textbf{0.77} \pm 0.02$  & $\textbf{1.00} \pm 0.00$    & $\textbf{0.62} \pm 0.02$  & $\textbf{0.71} \pm 0.01$  & $418.8 \pm \phantom 04.5$     & $\textbf{0.81} \pm 0.05$    \\
$512$h    & $0.72 \pm 0.03$           & $0.97 \pm 0.04$             & $0.58 \pm 0.06$           & $0.66 \pm 0.04$           & $\textbf{412.9} \pm 25.0$     & $0.67 \pm 0.15$ \\\hline
\end{tabular}
\end{table}

On the one hand, as the training and validation subset increases in size, the precision value improves but on the other hand recall value decreases as the subset grows, though at a smaller scale compared to the increase in precision.
This can be attributed to the fact that smaller subset sizes lead to a small validation set and therefore less data to obtain a threshold from. 
With a limited amount of data \lc{the validation set distribution is very different to the true data distribution}, leading to a threshold that is very small and hence marks most anomalies correctly but also leads to a lot of false positives. 
\lc{Despite the decreasing recall, the $F_1$ score increases with a growing training and validation subset size.
It can also be observed that both the $F_{1,\text{best}}$ and the $A_{\text{PR}}$ reach a point of diminishing returns after $8$h of dynamic testing. 
Given that neither metric relies on the unsupervised threshold, it indicates that it plays a large role in the $F_1$ score.
It further implies that the model quality remains largely the same from $8$h onwards.}
Also, the $F_1$ score seems to approach the $F_{1,\text{best}}$ score as the subset grows, also backing the fact that with a small subset size, a good threshold cannot easily be obtained.
Therefore, for application on the test benches in automotive powertrain development, the largest subset size is desirable due to the higher precision value and a \textit{closer-to-ideal} threshold value.
\lc{When using the unsupervised threshold, the average detection delay increases with a larger training and validation subset and appears to have a relationship with the recall figure.
This is due to the fact that $N_\text{tp}$ increases and $N_\text{fn}$ decreases accordingly, which means that the maximum delay $\delta=T$ is applied more often, leading to the larger average detection delay.
Lastly, the root-cause precision $P_\text{rc}$ also appears to have a positive correlation with the precision figure.
This is to be expected since a root-cause true positive can only occur when an anomaly is correctly identified. However, there is still clearly some room for improvement as $P_\text{rc}$ is always lower than or equal to $P$, where in case of a perfect root-cause analysis $P_\text{rc}=P$.}

\subsection{Reverse-window Process}
To investigate the effect of the mean-type reverse-window method, it is compared with the first-type and last-type methods where the first and last values of each window are carried over, respectively\lc{, the results for which are shown in Table \ref{tab:reverse}.}
\begin{table}[h!]
\centering
\caption{\lc{$F_1$ score, precision $P$, recall $R$, average detection delay $\bar{\delta}$ and root-cause precision $P_\text{rc}$ using the unsupervised threshold (top half) and theoretical best threshold (bottom half), as well as the area under the precision-recall curve $A_\text{PR}$ for the different reverse-window types. The best values for each metric are given in \textbf{bold}. The standard deviation for the different seeds are also provided.}}\label{tab:reverse}
\setlength{\tabcolsep}{3pt}
\begin{tabular}{lcccccc}
Type      & $F_1$                     & $P$                         & $R$                       & $A_{\text{PR}}$          & $\bar{\delta} \ [s]$              & $P_\text{rc}$   \\ \hline\hline
first     & $0.71 \pm 0.04$           & $\textbf{0.99} \pm 0.02$    & $0.55 \pm 0.05$           & $\textbf{0.68} \pm 0.06$  & $517.1 \pm \phantom 06.1$         & $\textbf{0.77} \pm 0.07$   \\
last      & $\textbf{0.72} \pm 0.01$  & $0.94 \pm 0.02$             & $\textbf{0.58} \pm 0.01$  & $0.64 \pm 0.01$           & $\textbf{411.9} \pm 15.7$         & $0.64 \pm 0.06$   \\
mean      & $0.70 \pm 0.03$           & $0.92 \pm 0.08$             & $0.57 \pm 0.04$           & $0.66 \pm 0.04$           & $418.1 \pm 17.3$                  & $0.63 \pm 0.18$ \\ \hline
first     & $\textbf{0.75} \pm 0.06$  & $0.95 \pm 0.04$             & $\textbf{0.62} \pm 0.10$  & $\textbf{0.68} \pm 0.06$  & $464.8 \pm 38.7$                  & $\textbf{0.70} \pm 0.10$   \\
last      & $0.72 \pm 0.01$           & $0.94 \pm 0.02$             & $0.58 \pm 0.01$           & $0.64 \pm 0.01$           & $\textbf{411.9} \pm 15.7$         & $0.64 \pm 0.06$   \\
mean      & $0.72 \pm 0.03$           & $\textbf{0.97} \pm 0.04$    & $0.58 \pm 0.06$           & $0.66 \pm 0.04$           & $412.9 \pm 25.0$                  & $0.67 \pm 0.15$ \\\hline
\end{tabular}
\end{table}

\lc{As is evident, the calibrated and uncalibrated detection performance is very similar throughout the different methods. 
In terms of average detection delay, it is evident, however, that the first-type method is significantly slower than the other two methods. 
Interestingly, despite the capacity to detect anomalies with much lower theoretical delay for most of the time steps in the sequence, the last-type actually yielded very similar average detection delays to the mean-type.
The root-cause precision is very similar for last and mean-type reverse-windowing, with first-type scoring the highest.
Furthermore,} the mean-type reverse-window method results in a higher computational load, though negligible. 

\subsection{Hyperparameter Optimisation}
As part of the hyperparameter optimisation of TeVAE, a list of key dimension sizes $d_\textbf{K}$ in combination with a list of latent dimension sizes $d_\textbf{Z}$ is tested. 
\lc{Note that, given the unsupervised nature of the problem, this optimisation is not possible in a productive environment. 
The results depict theoretical and in reality unobservable anomaly detection performance.
Alternatively, optimising for another metric like validation loss would be possible, however, there is no guarantee that said metric leads to good anomaly detection performance.}
Despite the larger learning capacity associated with a higher $d_\textbf{K}$, the \lc{attention head} concatenation is always transformed to an \lc{output} matrix of \lc{dimensionality} $d_\textbf{O}=d_\textbf{Z}$. 
\lc{For the two variables, values of 1, 8, 64, and 512 are tested; the results are shown in Tables \ref{tab:hyperparameter_dk} and \ref{tab:hyperparameter_dz}.}
\begin{table}[h!]
\caption{\lc{$F_1$ score, precision $P$, recall $R$, average detection delay $\bar{\delta}$ and root-cause precision $P_\text{rc}$ using the unsupervised threshold (top half) and theoretical best threshold (bottom half), as well as the area under the precision-recall curve $A_\text{PR}$ for various $d_\textbf{K}$ values.}}\label{tab:hyperparameter_dk}
\centering
\setlength{\tabcolsep}{3pt}
\begin{tabular}{cccccccc}
$d_\mathbf{K}$ & $d_\mathbf{Z}$  & $F_1$                     & $P$                      & $R$                       & $A_{\text{PR}}$           & $\bar{\delta} \ [s]$                  & $P_\text{rc}$   \\ \hline\hline
$1$            & $64$            & $\textbf{0.70} \pm 0.03$  & $0.92 \pm 0.08$          & $\textbf{0.57} \pm 0.04$  & $\textbf{0.66} \pm 0.04$  & $\textbf{418.1} \pm 17.3$             & $0.63 \pm 0.18$ \\
$8$            & $64$            & $0.69 \pm 0.01$           & $0.98 \pm 0.03$          & $0.54 \pm 0.02$           & $0.65 \pm 0.02$           & $446.7 \pm 36.9$                      & $0.66 \pm 0.06$ \\
$64$           & $64$            & $0.69 \pm 0.03$           & $\textbf{0.99} \pm 0.02$ & $0.53 \pm 0.03$           & $0.64 \pm 0.02$           & $457.9 \pm 18.3$                      & $0.71 \pm 0.01$ \\
$512$          & $64$            & $0.66 \pm 0.06$           & $\textbf{0.99} \pm 0.02$ & $0.50 \pm 0.07$           & $0.63 \pm 0.03$           & $498.9 \pm 41.6$                      & $\textbf{0.78} \pm 0.01$ \\\hline
$1$            & $64$            & $\textbf{0.72} \pm 0.03$  & $0.97 \pm 0.04$          & $\textbf{0.58} \pm 0.06$  & $\textbf{0.66} \pm 0.04$  & $412.9 \pm 25.0$                      & $0.67 \pm 0.15$ \\
$8$            & $64$            & $0.71 \pm 0.01$           & $0.92 \pm 0.08$          & $\textbf{0.58} \pm 0.04$  & $0.65 \pm 0.02$           & $\textbf{411.7} \pm \phantom 05.3$    & $0.62 \pm 0.02$ \\
$64$           & $64$            & $0.71 \pm 0.02$           & $0.98 \pm 0.03$          & $0.55 \pm 0.02$           & $0.64 \pm 0.02$           & $455.4 \pm 12.8$                      & $0.71 \pm 0.02$ \\
$512$          & $64$            & $0.70 \pm 0.03$           & $\textbf{0.99} \pm 0.02$ & $0.54 \pm 0.04$           & $0.63 \pm 0.03$           & $478.5 \pm 18.4$                      & $\textbf{0.75} \pm 0.05$\\\hline
\end{tabular}
\end{table}

\begin{table}[h!]
\caption{\lc{$F_1$ score, precision $P$, recall $R$, average detection delay $\bar{\delta}$ and root-cause precision $P_\text{rc}$ using the unsupervised threshold (top half) and theoretical best threshold (bottom half), as well as the area under the precision-recall curve $A_\text{PR}$ for various $d_\textbf{Z}$ values.}}\label{tab:hyperparameter_dz}
\centering
\setlength{\tabcolsep}{3pt}
\begin{tabular}{cccccccc}
$d_\mathbf{K}$ & $d_\mathbf{Z}$  & $F_1$                     & $P$                      & $R$                       & $A_{\text{PR}}$           & $\bar{\delta} \ [s]$      & $P_\text{rc}$   \\ \hline\hline
$1$            & $1$             & $0.34 \pm 0.18$           & $\textbf{1.00} \pm 0.00$ & $0.22 \pm 0.15$           & $0.42 \pm 0.11$           & $690.5 \pm 106.8$         & $0.74 \pm 0.15$ \\
$1$            & $8$             & $\textbf{0.75} \pm 0.03$  & $0.95 \pm 0.04$          & $\textbf{0.62} \pm 0.06$  & $\textbf{0.71} \pm 0.05$  & $\textbf{407.8} \pm 24.1$ & $0.55 \pm 0.10$ \\
$1$            & $64$            & $0.70 \pm 0.03$           & $0.92 \pm 0.08$          & $0.57 \pm 0.04$           & $0.66 \pm 0.04$           & $418.1 \pm 17.3$          & $0.63 \pm 0.18$ \\
$1$            & $512$           & $0.70 \pm 0.05$           & $0.98 \pm 0.03$          & $0.55 \pm 0.05$           & $0.64 \pm 0.03$           & $448.9 \pm 38.0$          & $\textbf{0.67} \pm 0.08$ \\ \hline
$1$            & $1$             & $0.49 \pm 0.10$           & $0.74 \pm 0.29$          & $0.44 \pm 0.12$           & $0.42 \pm 0.11$           & $655.3 \pm 99.8$          & $0.58 \pm 0.27$ \\
$1$            & $8$             & $\textbf{0.77} \pm 0.03$  & $0.94 \pm 0.05$          & $\textbf{0.65} \pm 0.06$  & $\textbf{0.71} \pm 0.05$  & $\textbf{387.6} \pm 37.6$ & $0.48 \pm 0.08$ \\
$1$            & $64$            & $0.72 \pm 0.03$           & $0.97 \pm 0.04$          & $0.58 \pm 0.06$           & $0.66 \pm 0.04$           & $412.9 \pm 25.0$          & $\textbf{0.67} \pm 0.15$ \\
$1$            & $512$           & $0.72 \pm 0.03$           & $\textbf{1.00} \pm 0.00$ & $0.56 \pm 0.04$           & $0.64 \pm 0.03$           & $439.7 \pm 24.2$          & $0.66 \pm 0.04$ \\\hline
\end{tabular}
\end{table}
\lc{As shown in the ablation study, the multi-head attention mechanism does positively impact anomaly detection performance, however Table \ref{tab:hyperparameter_dk} illustrates that once MA is implemented, the key dimensionality plays a small role, as all metrics are very comparable for all $d_\textbf{K}$.
When it comes to the latent dimension size $d_\textbf{Z}$, it is clear that for $d_\textbf{Z}=d_\textbf{K}=1$, the model cannot pass enough information through the two bottlenecks to yield good performance. 
Once $d_\textbf{Z}=8$ is reached, the best anomaly detection performance ever observed in the experimentation is obtained, after which it drops slightly for $d_\textbf{Z}=64$ and $d_\textbf{Z}=512$.
The same cannot be said for the root-cause precision $P_\text{rc}$, which is lower for $d_\textbf{Z}=8$ than for the other configurations.}

\subsection{Benchmarking}
Of course, TeVAE is not the first model proposed for time-series anomaly detection. 
To relate its anomaly detection performance, it is compared with a series of other models based on variational autoencoders. 
The chosen subset of models is based on the work discussed in Section \ref{sec:related_work} which either linked source code or contained enough information for implementation. 
The models are implemented using hyperparameters specified in their respective publications. 
All models are trained on the $512$h subset with early stopping, which is parameterised equally across all models. 
The anomaly detection process specified in Algorithm \ref{alg:anomaly_detection} is also applied to all models, along with the threshold estimation method. 
\lc{For VASP and LW-VAE no root-cause precision $P_\text{rc}$ is provided because the resulting anomaly scores cannot be decomposed.}
The results can be seen in Table \ref{tab:benchmark}.
\begin{table}[h!]
\caption{\lc{$F_1$ score, precision $P$, recall $R$, average detection delay $\bar{\delta}$ and root-cause precision $P_\text{rc}$ using the unsupervised threshold (top half) and theoretical best threshold (bottom half), as well as the area under the precision-recall curve $A_\text{PR}$ for competing models and TeVAE (Ours). The best values for each metric are given in \textbf{bold}.}}\label{tab:benchmark}
\centering
\setlength{\tabcolsep}{3pt}
\begin{tabular}{lcccccc}
Model   & $F_1$                     & $P$                       & $R$                       & $A_{\text{PR}}$           & $\bar{\delta} \ [s]$          & $P_\text{rc}$    \\ \hline\hline
VS-VAE  & $0.58 \pm 0.07$           & $0.90 \pm 0.10$           & $0.44 \pm 0.09$           & $0.56 \pm 0.06$           & $539.3 \pm 55.0$              & $0.71 \pm 0.17$  \\
OmniA   & $0.36 \pm 0.25$           & $0.65 \pm 0.46$           & $0.25 \pm 0.17$           & $0.39 \pm 0.14$           & $729.0 \pm 100.9$             & $0.61 \pm 0.43$  \\
W-VAE   & $0.46 \pm 0.02$           & $0.86 \pm 0.10$           & $0.31 \pm 0.01$           & $0.42 \pm 0.04$           & $596.9 \pm 12.8$              & $\textbf{0.86} \pm 0.10$  \\
SISVAE  & $0.35 \pm 0.22$           & $0.89 \pm 0.08$           & $0.25 \pm 0.16$           & $0.48 \pm 0.04$           & $665.4 \pm 100.5$             & $0.51 \pm 0.36$  \\ 
VASP    & $0.51 \pm 0.02$           & $\textbf{0.94} \pm 0.00$  & $0.35 \pm 0.02$           & $0.59 \pm 0.03$           & $581.6 \pm \phantom 03.5$     & n/a  \\ 
LW-VAE  & $0.48 \pm 0.01$           & $\textbf{0.94} \pm 0.00$  & $0.33 \pm 0.01$           & $0.57 \pm 0.03$           & $585.1 \pm \phantom 02.0$     & n/a  \\ 
TeVAE   & $\textbf{0.70} \pm 0.03$  & $0.92 \pm 0.08$           & $\textbf{0.57} \pm 0.04$  & $\textbf{0.66} \pm 0.04$  & $\textbf{418.1} \pm 17.3$     & $0.76 \pm 0.08$ \\ \hline
VS-VAE  & $0.62 \pm 0.07$           & $\textbf{0.97} \pm 0.02$  & $0.45 \pm 0.07$           & $0.56 \pm 0.06$           & $531.3 \pm 39.8$              & $0.75 \pm 0.10$  \\
OmniA   & $0.49 \pm 0.12$           & $0.72 \pm 0.37$           & $0.50 \pm 0.14$           & $0.39 \pm 0.14$           & $629.6 \pm 116.3$             & $0.62 \pm 0.35$  \\
W-VAE   & $0.50 \pm 0.04$           & $0.83 \pm 0.12$           & $0.36 \pm 0.04$           & $0.42 \pm 0.04$           & $576.0 \pm 37.1$              & $0.77 \pm 0.13$  \\
SISVAE  & $0.55 \pm 0.04$           & $0.95 \pm 0.01$           & $0.39 \pm 0.04$           & $0.48 \pm 0.04$           & $599.4 \pm 11.0$              & $\textbf{0.88} \pm 0.03$  \\ 
VASP    & $0.66 \pm 0.03$           & $0.80 \pm 0.03$           & $0.57 \pm 0.07$           & $0.59 \pm 0.03$           & $474.6 \pm 20.1$              & n/a  \\ 
LW-VAE  & $0.64 \pm 0.02$           & $0.78 \pm 0.03$           & $0.54 \pm 0.04$           & $0.57 \pm 0.03$           & $489.3 \pm 14.2$              & n/a  \\ 
TeVAE   & $\textbf{0.72} \pm 0.03$  & $\textbf{0.97} \pm 0.04$  & $\textbf{0.58} \pm 0.06$  & $\textbf{0.66} \pm 0.04$  & $\textbf{412.9} \pm 25.0$     & $0.80 \pm 0.09$ \\\hline
\end{tabular}
\end{table}

\lc{As is evident, TeVAE outperforms all other models in $F_1$ score, $R$, $A_{\text{PR}}$ and $\bar{\delta}$, while providing nearly matching the best precision result using the unsupervised threshold.}
As stated in Section \ref{sec:proposed_approach} a very high precision figure is important in this type of powertrain testing, however, the reduced precision is still considered tolerable. 
Also, it comes at the benefit of a much higher recall figure, which is reflected in the superior $F_1$ figure\lc{, though it still leaves room for improvement.}
Furthermore, the $F_{1,\text{best}}$ figure, which is obtained at $P=0.97$ and $R=0.58$, suggests that TeVAE has the potential to achieve even higher precision without sacrificing recall if the threshold were optimised. 

\section{Conclusion and Outlook}\label{sec:conclusion}
In this paper, a variational autoencoder (TeVAE) for unsupervised anomaly detection in automotive testing is proposed.
\lc{Automotive testing is an especially challenging scenario due to its massive, diverse and multi-dimensional nature.
In addition to that, the resulting data not only features variable states but also highly dynamic signals along with more static ones, adding to further complexity.}
It not only features an attention configuration that avoids the bypass phenomenon but also introduces a novel method of remapping windows to whole sequences.
A number of experiments are conducted to demonstrate the online anomaly detection performance of the model, as well as to underline the benefits of key aspects introduced with the model.
\lc{To this end, novel metrics are introduced to measure the detection delay, as well as the root-cause analysis capability of analysed anomaly detection approaches.}

From the results obtained, TeVAE clearly benefits from the MA mechanism, indicating the avoidance of the bypass phenomenon. 
Moreover, the proposed approach only requires a small training and validation subset size but fails to obtain a suitable threshold, as with increasing subset size only the calibrated anomaly detection performance increases.
\lc{Despite the higher theoretical delay, mean-type reverse windowing performs comparably to its last-type in both detection performance and observed average detection delay, while outperforming the first-type method, which in turn yielded higher root-cause precision than the other two methods.
Also, the hyperparameter optimisation reveals that one of the least parameter-heavy configurations of the TeVAE results in the best anomaly detection performance, though it should be noted that this performance is only theoretical and not achievable in a production environment due to the unsupervised nature of the problem.}
In its default setting, TeVAE is only \lc{$8\%$} of the time wrong when an anomaly is flagged and manages to discover \lc{$57\%$} of the anomalies present in the test data set when the proposed parameter-free and unsupervised threshold is used. 
\lc{If the ideal threshold were known, these values improve to $3\%$ and $58\%$, respectively.}
Lastly, it outperforms all other competing models it is compared with.

In the future, \lc{active learning will be investigated in the context of threshold choice in an effort to find a suitable threshold at an earlier stage.}

\printbibliography %Prints bibliography

\end{document}